\documentclass[11pt,a4paper]{article}
\usepackage[hyperref]{emnlp2020}
\usepackage{times}
\usepackage{latexsym}

\usepackage{algorithm}
\usepackage{algpseudocode}
\usepackage{amsmath}
\usepackage{tikz}
\usepackage{csquotes}
\usepackage{amssymb}
\usepackage{enumitem}
\usepackage{tablefootnote}
\usepackage{longtable}
\usepackage{pgfplots}
\usepackage{booktabs}
\usepackage{subcaption}
\usepackage{siunitx}
\usepackage{cleveref}
\usepackage{multirow}
\usepackage{dblfloatfix}
\crefformat{section}{\S#2#1#3} 
\crefformat{subsection}{\S#2#1#3}
\crefformat{subsubsection}{\S#2#1#3}
\usetikzlibrary{intersections,shapes.arrows}

\makeatletter
\let\OldStatex\Statex
\renewcommand{\Statex}[1][3]{%
  \setlength\@tempdima{\algorithmicindent}%
  \OldStatex\hskip\dimexpr#1\@tempdima\relax}
\makeatother

\algrenewcommand\algorithmicindent{1.3em}

\usepackage{microtype}

\aclfinalcopy 

\newcommand{\comment}[1]{}

\def\D{\mathcal{D}}

\def\ssmba{SSMBA}

\def\M{\mathcal{M}}

\title{SSMBA: Self-Supervised Manifold Based Data Augmentation for Improving Out-of-Domain Robustness}

\author{Nathan Ng\\
  University of Toronto \\
  Vector Institute
  \And
  Kyunghyun Cho \\
  New York University
  \And
  Marzyeh Ghassemi \\
  University of Toronto \\
  Vector Institute\\ 
  }

\date{}

\begin{document}
\maketitle
\begin{abstract}
  Models that perform well on a training domain often fail to generalize to out-of-domain (OOD) examples.
  Data augmentation is a common method used to prevent overfitting and improve OOD generalization.
  However, in natural language, it is difficult to generate new examples that stay on the underlying data manifold.
  We introduce \textbf{\ssmba}, a data augmentation method for generating synthetic training examples by using a pair of corruption and reconstruction functions to move randomly on a data manifold.
  We investigate the use of \ssmba\ in the natural language domain, leveraging the manifold assumption to reconstruct corrupted text with masked language models. 
  In experiments on robustness benchmarks across 3 tasks and 9 datasets, \ssmba\ consistently outperforms existing data augmentation methods and baseline models on both in-domain and OOD data, achieving gains of 0.8\% accuracy on OOD Amazon reviews, 1.8\% accuracy on OOD MNLI, and 1.4 BLEU on in-domain IWSLT14 German-English. 
  \footnote{Code is availble at \url{https://github.com/nng555/ssmba}}
\end{abstract}

\section{Introduction}
\label{sec:intro}
Training distributions often do not cover all of the test distributions we would like a supervised classifier or model to perform well on.
Often, this is caused by biased dataset collection \citep{torralba2011unbiased} or test distribution drift over time \citep{quionera2009dataset}.
Therefore, a key challenge in training machine learning models in these settings is ensuring they are robust to unseen examples. 
Since it is impossible to generalize to the entire distribution, methods often focus on the adjacent goal of \textit{out-of-domain robustness}.

Data augmentation is a common technique used to improve out-of-domain (OOD) robustness by synthetically generating new training examples \citep{Simard1998}, often by perturbing existing examples in the input space \citep{perez2017}. 
If data concentrates on a low-dimensional manifold \citep{chapelle2006semi}, then these synthetic examples should lie in a manifold neighborhood of the original examples \citep{vicinal200olivier}.
Training models to be robust to such local perturbations has been shown to be effective in improving performance and generalization in semi-supervised and self-supervised settings \citep{bachman2014learning,szegedy2014intriguing, sajjadi2016regularization}.
When the underlying data manifold exhibits easy-to-characterize properties, as in natural images, simple transformations such as translation and rotation 
can quickly generate local training examples.
However, in domains such as natural language, it is much more difficult to find a set of invariances that preserves meaning or semantics.

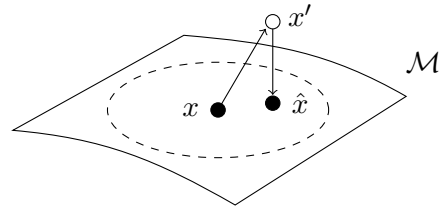
\begin{figure}[t!]
\centering
\begin{tikzpicture}[scale=0.9]

\draw (0,0) to[out=-5,in=150] (3.25,-1.1) -- (5.75,0.5) to[out=150,in=-5] (2.5,1.4) -- cycle;
\node at (6, 1) {$\M$};

\coordinate (x) at (3.0, 0.3);
\draw[fill] (x) circle (3pt);
\node at (2.6, 0.3) {$x$};

\coordinate (qx) at (3.8, 1.6);
\draw (qx) circle (3pt);
\node at (4.2, 1.7) {$x'$};

\coordinate(rqx) at (3.8, 0.4);
\draw[fill] (rqx) circle (3pt);
\node at (4.2, 0.4) {$\hat{x}$};

\draw [dashed] (x) ellipse (46pt and 20pt);

\draw [->] (x) -- ([xshift=-3pt,yshift=-3pt]qx);
\draw [->] ([yshift=-3pt]qx) -- ([yshift=3.5pt]rqx);
\end{tikzpicture}
\caption{\ssmba\ moves along the data manifold $\M$ by using a corruption function to perturb an example $x$ off the data manifold, then using 
a reconstruction function to project it back on.}
\label{fig:ssmba_process}
\end{figure}

In this paper we propose \textbf{S}elf-\textbf{S}upervised \textbf{M}anifold \textbf{B}ased Data \textbf{A}ugmentation (\textbf{\ssmba}): a data augmentation method for generating synthetic examples in domains where the data manifold is difficult to heuristically characterize.
Motivated by the use of denoising auto-encoders as generative models \citep{bengio2013generalized}, we use a corruption function to stochastically perturb examples \textit{off} the data manifold, then use a reconstruction function to project them \textit{back on} (Figure \ref{fig:ssmba_process}).
This ensures new examples lie within the manifold neighborhood of the original example.
SSMBA is applicable to any supervised task, requires no task-specific knowledge, and does not rely on class- or dataset-specific fine-tuning.

We investigate the use of \ssmba\ in the natural language domain on 3 diverse tasks spanning both classification and sequence modelling: sentiment analysis, natural language inference, and machine translation.
In experiments across 9 datasets and 4 model types, we show \ssmba\ consistently outperforms baseline models and other data augmentation methods on both in-domain and OOD data. 

\section{Background and Related Work}
\label{sec:background}
\subsection{Data Augmentation in NLP}
The problem of domain adaptation and OOD robustness is well established in NLP \citep{blitzer-etal-2007-biographies,daume-iii-2007-frustratingly,hendrycks2020pretrained}.
Existing work on improving generalization has focused on data augmentation, where synthetically generated training examples are used to augment an existing dataset.
It is hypothesized that these examples induce robustness to local perturbations, which has been shown to be effective in semi-supervised and self-supervised settings \citep{bachman2014learning,szegedy2014intriguing, sajjadi2016regularization}.

Existing task-specific methods \citep{kafle-etal-2017-data} and word-level methods \citep{zhang2015character, xie2017data, wei-zou-2019-eda} are based on human-designed heuristics.
Back-translation from or through another language has been applied in the context of machine translation \citep{sennrich2016improving}, question answering \citep{wei2018fast}, and consistency training \citep{xie2019unsupervised}.
More recent work has used word embeddings \citep{wangyang2015thats} and LSTM language models \citep{fadaee2017data} to perform word replacement.
Other methods focus on fine-tuning contextual language models \citep{kobayashi-2018-contextual,wu2019conditional,kumar20202data} or large generative models \citep{lambada,yang2020g-daug,kumar20202data} to generate synthetic examples.

\subsection{VRM and the Manifold Assumption}
Vicinal Risk Minimization (VRM) \citep{vicinal200olivier} formalizes data augmentation as enlarging the training set support by drawing samples from a \textit{vicinity} of existing training examples.
Typically the vicinity of a training example is defined using dataset-dependent heuristics.
For example, in computer vision, examples are generated using scale augmentation \citep{simonyan2014very}, color augmentation \citep{krizhevsky2012imagenet}, and translation and rotation \citep{Simard1998}.

The \textit{manifold assumption} states that high dimensional data concentrates around a low-dimensional manifold \citep{chapelle2006semi}.
This assumption allows us to define the vicinity of a training example as its \textit{manifold neighborhood}, the portion of the neighborhood that lies on the data manifold.
Recent methods have used the manifold assumption to improve robustness by moving examples towards a decision boundary \citep{kanbak2018geometric}, generating adversarial examples \cite{szegedy2014intriguing,miyato2017virtual}, interpolating between pairs of examples \citep{zhang2018mixup}, or finding affine transforms \citep{paschali2019data}.

\begin{figure}[t!]
\centering
\includegraphics[scale=0.21]{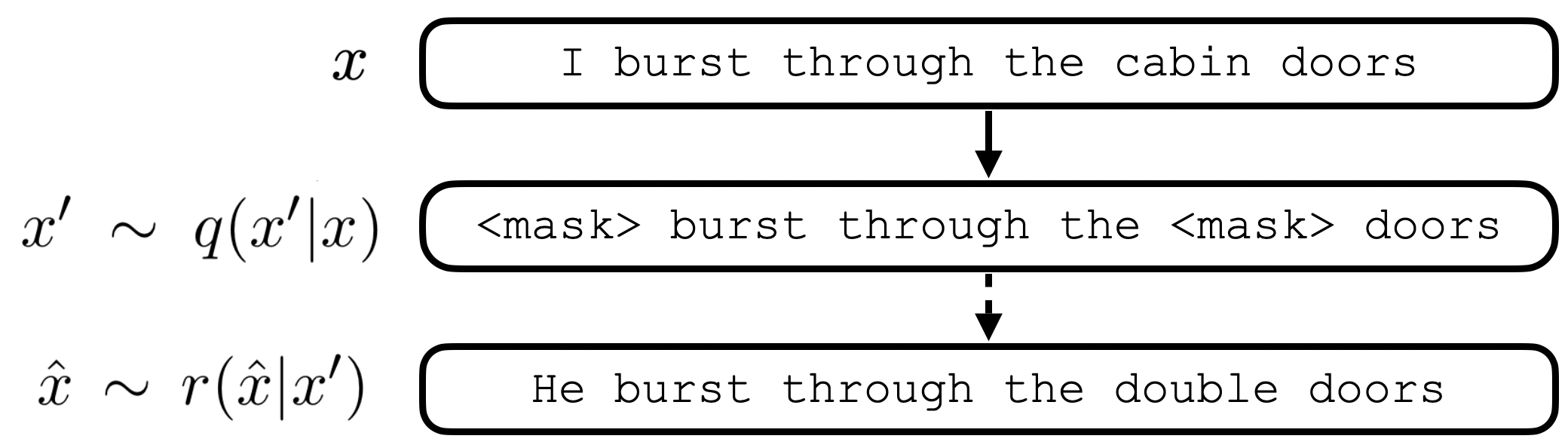}
\caption{To sample from an MLM DAE, we apply the MLM corruption $q$ to the original sentence then reconstruct the corrupted sentence using our DAE $r$.}
\label{fig:dae_sampling}
\end{figure}

\subsection{Sampling from Denoising Autoencoders}
A denoising autoencoder (DAE) is an autoencoder trained to reconstruct a clean input $x$ from a stochastically corrupted one $x'\sim q(x'|x)$ by learning a conditional distribution $P_\theta (x| x')$ \citep{vincent2008extracting}.
We can sample from a DAE by successively corrupting and reconstructing an input using the following pseudo-Gibbs Markov chain: $x_t' \sim q(x'|x_{t-1})$, $x_t \sim P_\theta(x|x'_t).$
\comment{
\begin{align*}
    x_t' &\sim q(x'|x_{t-1})\\
    x_t &\sim P_\theta(x|x'_t) 
\end{align*}
}
As the number of training examples increases, the asymptotic distribution $\pi_n(x)$ of the generated samples approximate the true data-generating distribution $P(x)$ \citep{bengio2013generalized}.
This corruption-reconstruction process allows for sampling directly along the manifold that $P(x)$ concentrates on.

\subsection{Masked Language Models}
Recent advances in unsupervised representation learning for natural language have relied on pre-training models on a \textit{masked language modeling} (MLM) objective \citep{devlin2018, liu2019roberta}.
In the MLM objective, a percentage of the input tokens are randomly corrupted and the model is asked to reconstruct the original token given its left and right context in the corrupted sentence.
We use MLMs as DAEs \citep{lewis2019bart} to sample from the underlying natural language distribution by corrupting and reconstructing inputs (Figure \ref{fig:dae_sampling}).

\section{SSMBA: Self-Supervised Manifold Based Augmentation}
\label{sec:ssmba}
\begin{algorithm}
\begin{algorithmic}[1]
    \State \textbf{Require:} \parbox[t]{\dimexpr\linewidth-\algorithmicindent}{perturbation function $q$\\ reconstruction function $r$ \strut}
    \State \textbf{Input:} \parbox[t]{\dimexpr\linewidth-\algorithmicindent}{Dataset $\D = \{(x_1, y_1)\ldots(x_n, y_n)\}$\\ number of augmented examples $m$ \strut}
    \Function{SSMBA}{$\D$, $m$}
    \State train a model $f$ on $\D$
    \For{$(x_i, y_i) \in \D$}
        \For{$j \in 1\ldots m$}
            \State sample perturbed $x_{ij}' \sim q(x'|x_i)$
            \State sample reconstructed $\hat{x}_{ij} \sim r(\hat{x}|x_{ij}')$ 
            \State generate $\hat{y}_{ij} \gets f(\hat{x}_{ij})$ or preserve \Statex[3] the original $y_i$
        \EndFor
    \EndFor
    \State let $\D^{aug} = \{(\hat{x}_{ij}, \hat{y}_{ij})\}_{i=1\ldots n,j=1\ldots m}$
    \State augment $ \D' \leftarrow \D \cup \D^{aug}$ 
    \State \Return $\D'$
    \EndFunction
\end{algorithmic}
\caption{SSMBA}
\label{alg:ssmba}
\end{algorithm}

\begin{figure}[t]
\centering
\includegraphics[scale=0.225]{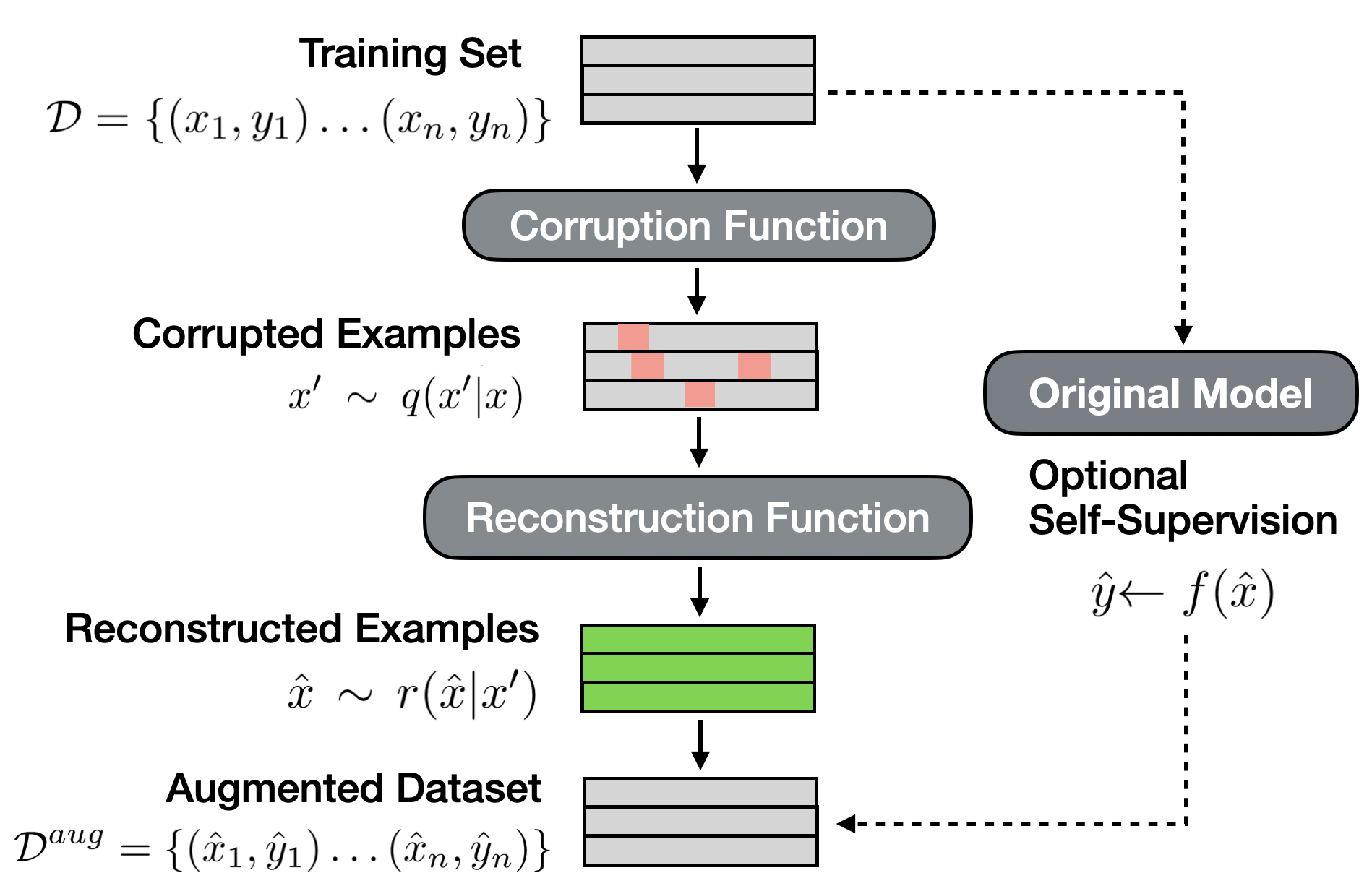}
\caption{\ssmba\ generates synthetic examples by corrupting then reconstructing the original training inputs. 
To form the augmented dataset, corresponding outputs are preserved from the original data or generated from a supervised model $f$ trained on the original data.}
\label{fig:ssmba_graph}
\end{figure}

\noindent We now describe \textbf{S}elf-\textbf{S}upervised \textbf{M}anifold \textbf{B}ased Data \textbf{A}ugmentation. 
Let our original dataset $\D$ consist of pairs of input and output vectors $\D = \{(x_1, y_1)\ldots(x_n, y_n)\}$.
We assume the input points concentrate around an underlying lower dimensional data manifold $\M$.
Let $q$ be a corruption function from which we can draw a sample $x' \sim q(x'|x)$ such that $x'$ no longer lies on $\M$.
Let $r$ be a reconstruction function from which we can draw a sample $\hat{x} \sim r(\hat{x}|x')$ such that $\hat{x}$ lies on $\M$. 

To generate an augmented dataset, we take
each pair $(x_i, y_i)\in\D$ and sample a perturbed $x_i' \sim q(x'|x_i)$.
We then sample a reconstructed $\hat{x}_{ij} \sim r(\hat{x}|x_i')$.
A corresponding vector $\hat{y}_{ij}$ can be generated by preserving $y_i$, or, 
since examples in the manifold neighborhood may cross decision boundaries on more sensitive tasks, by using a teacher model trained on the original data.
This operation can be repeated to generate multiple augmented examples for each input example.
These new examples form a dataset that we can augment the original training set with. 
We can then train an augmented model on the new augmented dataset.

In this paper we investigate \ssmba's use on natural language tasks, using the MLM training corruption function as our corruption function $q$ and a pre-trained BERT model as our reconstruction model $r$.
Different from other data augmentation methods, 
\ssmba\ does not rely on task-specific knowledge, requires no dataset-specific fine-tuning, and is applicable to any supervised natural language task.
\ssmba\ requires only a pair of functions $q$ and $r$ used to generate data.

\comment{
\begin{figure*}[t!]
\centering
\begin{tikzpicture}

\draw (0,0) to[out=-5,in=150] (5,-1) -- (7,1) to[out=150,in=-10] (2.7,2.0) -- cycle;
\node at (6, 1) {$\M$};

\coordinate (x) at (2, 0.5);
\draw[fill] (x) circle (3pt);
\node at (1.6, 0.5) {$x$};

\coordinate (qx) at (3, 1.4);
\draw (qx) circle (3pt);
\node at (4, 1.4) {$x' \sim q(x)$};

\coordinate(rqx) at (2.9, 0.2);
\draw[fill=gray] (rqx) circle (3pt);
\node at (4, 0.2) {$\hat{x} \sim r(x')$};

\draw [->] (x) -- ([xshift=-3pt,yshift=-3pt]qx);
\draw [dashed, ->] ([yshift=-3pt]qx) -- ([yshift=3.5pt]rqx);
\end{tikzpicture}
\caption{Noising and reconstruction process for a given point $x$ lying on a data manifold $\M$}
\end{figure*}
}

\section{Datasets}
\label{sec:data}
To empirically evaluate our proposed algorithm, we select 9 datasets -- 4 sentiment analysis datasets, 2 natural language inference (NLI) datasets, and 3 machine translation (MT) datasets.
Table \ref{tab:data_summ_small} and Appendix \ref{app:data} provide dataset summary statistics.
All datasets either contain metadata that can be used to split the samples into separate domains or similar datasets that are treated as separate domains. 

\subsection{Sentiment Analysis}        
The Amazon Review Dataset \citep{jianmo} contains product reviews from Amazon. 
Following \citealt{hendrycks2020pretrained}, we form two datasets: 
\textbf{AR-Full} contains reviews from the 10 largest categories, and 
\textbf{AR-Clothing} contains reviews in the clothing category separated into subcategories by metadata. 
Since the reviews in AR-Clothing come from the same top-level category, the amount of domain shift is much less than that of AR-Full.
Models predict a review's 1 to 5 star rating.
    
SST2 \citep{socher2013recursive} contains movie review excerpts. 
Following \citealt{hendrycks2020pretrained} we pair this dataset with the IMDb dataset \citep{maas2011learning}, which contains full length movie reviews.
We call this pair the \textbf{Movies} dataset.
Models predict a movie review's binary sentiment. 
    
The \textbf{Yelp Review Dataset} contains restaurant reviews with associated business metadata which we preprocess following \citealt{hendrycks2020pretrained}.
Models predict a review's 1 to 5 star rating.

\begin{table}[t]
\small
\centering
\begin{tabular}{llllll}
\toprule
\textbf{Dataset} & \textbf{Domain} & \pmb{$n$} & \pmb{$l$} & \textbf{Train} & \textbf{Test} \\
\toprule
AR-Clothing & * & 4 & 35 & 25k$^\dagger$ & 2k \\
\midrule
AR-Full & * & 10 & 67 & 25k$^\dagger$ & 2k \\
\midrule
Yelp & * &  4 & 138 & 25k$^\dagger$ & 2k \\
\midrule
\multirow{2}{*}{Movies}
& SST2 &-  & 11 & 66k & 1k \\
& IMDb & - & 296 & 46k & 2k \\
\midrule
MNLI & *  & 10 & 36 & 80k & 1k \\
\midrule
\multirow{3}{*}{ANLI}
& R1 & - & 92 & 17k & 1k \\
& R2 & -  & 90 & 46k & 1k \\
& R3 & -  & 82 & 100k & 1k \\
\midrule
IWSLT & - & 1 & 24 & 160k & 7k \\
\midrule
OPUS & Medical & 5 & 15 & 1.1m & 2k \\
\midrule
\multirow{2}{*}{de-rm}
& Law & - & 22 & 100k & 2k\\
& Blogs & - & 25 & - & 2k \\
\bottomrule
\end{tabular}
\caption{Dataset summary statistics. $n$: number of domains. $l$: average tokenized input length. A * in the domain column indicates that the statistics are identical across domains within that dataset. Training sets marked with a $\dagger$ are sampled randomly from a larger dataset. Refer to Appendix \ref{app:data} for more information.}
\label{tab:data_summ_small}
\end{table}

\subsection{Natural Language Inference}
\textbf{MNLI} \citep{williams2018broad} is a corpus of NLI data from 10 distinct genres of written and spoken English.
We train on the 5 genres with training data and test on all 10 genres.
Since the dataset does not include labeled test data, we use the validation set as our test set and sample 2000 examples from each training set for validation.

\textbf{ANLI} \citep{nie2019adversarial} is a corpus of NLI data designed adversarially by humans such that state-of-the-art models fail to classify examples correctly.
The dataset consists of three different levels of difficulty which we treat as separate textual domains.

\subsection{Machine Translation}
Following \citealt{Muller2019DomainRI}, we consider two translation directions, German$\to$English (de$\to$en) and German$\to$Romansh (de$\to$rm). Romansh is a low-resource language with an estimated 40,000 native speakers where OOD robustness is of practical relevance \citep{Muller2019DomainRI}.

In the de$\to$en direction, we use \textbf{IWSLT14 de$\to$en} \citep{cettolo2014proceedings} as a widely-used benchmark to test in-domain performance. 
We also use the \textbf{OPUS} \citep{TIEDEMANN12.463} dataset to test OOD generalization. 
We train on highly specific in-domain data (medical texts) and disparate out-of-domain data (Koran text, Ubuntu localization files, movie subtitles, and legal text).
Since domains share very little similarities in language, generalization to out-of-domain text is extremely difficult.
In the \textbf{de$\to$rm} direction, we use a training set consisting of the Allegra corpus \citep{scherrer-cartoni-2012-trilingual} and Swiss press releases. We use blog posts from Convivenza as a test domain. 

\section{Experimental Setup}
\label{sec:experiments}
\subsection{Model Types}
For sentiment analysis tasks, we investigate LSTMs \citep{hochreiter1997long} and convolutional neural networks (CNNs).
For NLI tasks, we investigate fine-tuned RoBERTa\textsubscript{BASE} models \citep{liu2019roberta}, which are pretrained bidirectional transformers \citep{vaswani2017attention}.
On both tasks, representations from the encoder are fed into an feed-forward neural network for classification.
For MT tasks, we train transformers \citep{vaswani2017attention}.
For all models, word embeddings are initialized randomly and trained end-to-end with the model.
We do not initialize with pre-trained word embeddings to maintain consistency across all models and tasks.
Model hyperparameters are tuned to maximize performance on in-domain validation data. 
Training details and hyperparameters for all models are provided in Appendix \ref{app:hyperparams}.

\subsection{\ssmba\ Settings}
For all experiments we use the MLM corruption function as our corruption function $q$.
We tune tune the total percentage of tokens corrupted, leaving the percentages of specific corruption operations (80\% masked, 10\% random, 10\% unmasked) the same.
For sentiment analysis and NLI experiments we use a pre-trained RoBERTa\textsubscript{BASE} model as our reconstruction function $r$, and for translation experiments we use a pre-trained German BERT model \citep{chan2020deepset}.
For each input example, we generate 5 augmented examples using unrestricted sampling.
For translation experiments, target side translations are generated with beam search with width 5.
\ssmba\ hyperparameters, including augmented example labelling method and corruption percentage, are chosen based on in-domain validation performance.
Hyperparameters for each dataset are provided in Appendix \ref{app:ssmba}.


\begin{table*}[t!]
\small
\centering
\begin{tabular}{llcccccccccc}
\toprule
& & \multicolumn{2}{c}{\textbf{AR-Full}} & \multicolumn{2}{c}{\textbf{AR-Clothing}} & \multicolumn{2}{c}{\textbf{Movies}} & \multicolumn{2}{c}{\textbf{Yelp}} & \multicolumn{2}{c}{\textbf{Average}}\\
\cmidrule(lr){3-4}
\cmidrule(lr){5-6}
\cmidrule(lr){7-8}
\cmidrule(lr){9-10}
\cmidrule(lr){11-12}
Model & Augmentation & ID & OOD & ID & OOD & ID & OOD & ID & OOD & ID & OOD\\
\midrule
\multirow{5}{*}{RNN}
& None & 69.46 & 66.32 & 69.25 & 67.80 & 90.74 & 71.94 & 62.51 & 61.28 & 70.16 & 66.17 \\
& EDA & 67.32 & 64.47 & 66.87 & 65.21 & 88.43 & 68.3 & 58.39 & 57.19 & 67.56 & 63.55 \\
& CBERT & 69.94 & 66.77 & 69.56 & 68.10 & \textbf{91.01} & 72.11 & 63.17 & 61.75 & 70.17 & 66.57 \\
& UDA & 69.92 & 66.97 & 69.98 & 68.24 & 90.05 & 69.73 & 63.40 & 62.13 & 70.64 & 66.53 \\
\cmidrule{2-12}
& \ssmba & \textbf{70.38\rlap{$^{*\dagger}$} } & \textbf{67.41\rlap{$^{*\dagger}$}} & \textbf{70.19} & \textbf{68.60\rlap{$^{*\dagger}$}} & 89.61 & \textbf{73.20} & \textbf{63.85} & \textbf{62.83\rlap{$^{*\dagger}$}} & \textbf{70.96} & \textbf{67.31} \\
\midrule
\multirow{5}{*}{CNN}
& None & 70.67 & 67.64 & 70.14 & 68.52 & 92.92 & 72.11 & 65.13 & 64.46 & 71.68 & 67.63 \\
& EDA & 68.52 & 66.03 & 67.76 & 66.17 & 91.22 & 74.20 & 60.99 & 59.88 & 69.13 & 65.65 \\
& CBERT & 70.62 & 67.70 & 70.13 & 68.23 & 92.92 & 71.56 & 65.09 & 64.19 & 71.65 & 67.49 \\
& UDA & 70.80 & 68.06 & 70.29 & 68.70 & 92.63 & 72.55 & 65.22 & 64.32 & 71.77 & 67.89 \\
\cmidrule{2-12}
& \ssmba & \textbf{71.10\rlap{$^*$} } & \textbf{68.18\rlap{$^*$} } & \textbf{70.74} & \textbf{69.04\rlap{$^*$} } & \textbf{92.93} & \textbf{74.67} & \textbf{65.59} & \textbf{64.81\rlap{$^{*\dagger}$}} & \textbf{72.11} & \textbf{68.33} \\
\bottomrule
\end{tabular}
\caption{Average in-domain (ID) and out-of-domain (OOD) accuracy (\%) for models trained on sentiment analysis datasets. Average performance across datasets is weighted by number of domains contained in each dataset. Accuracies marked with a $*$ and $\dagger$ are statistically significantly higher than unaugmented models and the next best model respectively, both with $p<0.01$.}
\label{tab:sent_results}
\end{table*}


\subsection{Baselines}
On sentiment analysis and NLI tasks, we compare against 3 data augmentation methods.
Easy Data Augmentation (EDA) \citep{wei-zou-2019-eda} is a heuristic method that randomly replaces synonyms and inserts, swaps, and deletes words.
Conditional Bert Contextual Augmentation (CBERT) \citep{wu2019conditional} finetunes a class-conditional BERT model and uses it to generate sentences in a process similar to our own.
Unsupervised Data Augmentation (UDA) \citep{xie2020unsupervised} translates data to and from a pivot language to generate paraphrases. We adapt UDA for supervised classification tasks by training directly on the backtranslated data.

On translation tasks, we compare only against methods which do not require additional target side monolingual data.
Word dropout \citep{sennrich-etal-2016-edinburgh} randomly chooses words in the source sentence to set to zero embeddings.
Reward Augmented Maximum Likelihood (RAML) \citep{norouzi2016reward} samples noisy target sentences based on an exponential of their Hamming distance from the original sentence.
SwitchOut \citep{wang-etal-2018-switchout} applies a noise function similar to RAML to both the source and target side.
We use publicly available implementations for all methods. 

\subsection{Evaluation Method}
We train LSTM and CNN models with 10 random seeds, RoBERTa models with 5 random seeds, and transformer models with 3 random seeds.
Models are trained separately on each domain then evaluated on all domains, and performance is averaged across seeds and test domains.
We report the average in-domain (ID) and OOD performance across all train domains.
On sentiment analysis and NLI tasks we report accuracy, and on translation we report uncased tokenized BLEU \citep{papineni2002bleu} for IWSLT and cased, detokenized BLEU  with SacreBLEU\footnote{Signature: BLEU+c.mixed+\#1+s.exp+tok.13a+v.1.4.3} \citep{post-2018-call} for all others.
Statistical testing details are in Appendix \ref{app:stats}.

\section{Results}
\label{sec:results}

\subsection{Sentiment Analysis}
\label{subsec:sentiment_exp}
Table \ref{tab:sent_results} present results on sentiment analysis. 
Across all datasets, models trained with \ssmba\ outperform baseline models and all other data augmentation methods on OOD data.
On ID data, \ssmba\ outperforms baseline models and other data augmentation methods on all datasets for CNN models, and 3/4 datasets for RNN models.
On average, \ssmba\ improves OOD performance by 1.1\% for RNN models and 0.7\% for CNN models, and ID performance by 0.8\% for RNN models and 0.4\% for CNN model.
Other methods achieve much smaller OOD generalization gains and perform worse than baseline models on multiple datasets.

On the AR-Full dataset, RNNs trained with \ssmba\ demonstrate improvements in OOD accuracy of 1.1\% over baseline models.
On the AR-Clothing dataset, which exhibits less domain shift than AR-Full, RNNs trained with \ssmba\ exhibit slightly lower OOD improvement.
CNN models exhibit about the same boost in OOD accuracy across both Amazon review datasets.

On the Movies dataset where we observe a large difference in average sentence length between the two domains, \ssmba\ still manages to present considerable gains in OOD performance.
Although RNNs trained with \ssmba\ fail to improve ID performance, their OOD performance in this setting still beats other data augmentation methods.

On the Yelp dataset, we observe large performance gains on both ID and OOD data for RNN models.
The improvements on CNN models are more modest, but notably our method is the only one that improves OOD generalization.

\begin{table}[t!]
\small
\centering
\begin{tabular}{lcccc}
\toprule
& \multicolumn{2}{c}{\textbf{MNLI}} & \multicolumn{2}{c}{\textbf{ANLI}}\\
\cmidrule(lr){2-3}
\cmidrule(lr){4-5}
Augmentation  & ID & OOD & ID & OOD \\
\midrule
None & 84.29 & 80.61 & 42.54 & \textbf{43.80} \\
EDA & 83.44 & 80.34 & 45.59 & 42.77 \\
CBERT & 84.24 & 80.34 & 46.68 & 43.53 \\
UDA & 84.24 & 80.99 & 45.85 & 42.89 \\
\midrule
\ssmba\ & \textbf{85.71} & \textbf{82.44\rlap{$^{*\dagger}$}} & \textbf{48.46\rlap{$^{*\dagger}$}} & \textbf{43.80}  \\
\bottomrule
\end{tabular}
\caption{Average in-domain and out-of-domain accuracy (\%) for RoBERTa models trained on NLI tasks. Accuracies marked with a $*$ and $\dagger$
are statistically significantly higher than unaugmented models and the next best model respectively, both with $p<0.01$.}
\label{tab:nli_results}
\end{table}

\subsection{Natural Language Inference}
\label{subsec:nli_exp}
Table \ref{tab:nli_results} presents results on NLI tasks.
Models trained with \ssmba\ outperform or match baseline models and data augmentation methods on both ID and OOD data. 
Even with a more difficult task and stronger baseline model, \ssmba\ still confers large accuracy gains.
On MNLI, \ssmba\ improves OOD accuracy by 1.8\%, while
the best performing baseline achieves only 0.3\% improvement.
Our method also improves ID accuracy by 1.4\%.
All other baseline methods hurt both ID and OOD accuracy, or confer negligible improvements.

On the intentionally difficult ANLI, \ssmba\ maintains baseline OOD accuracy while conferring a large 6\% improvement on ID data. 
Other augmentation methods improve ID accuracy by a much smaller margin while degrading OOD accuracy.
Surprisingly, pseudo-labelling augmented examples in the R2 and R3 domains produced the best results, even when the labelling model had poor in-domain performance.

\begin{table}[t!]
\small
\centering
\begin{tabular}{lc}
\toprule
System & BLEU\\
\midrule
ConvS2S \citep{edunov-etal-2018-classical} & 32.2 \\
Transformer \citep{wu2018pay} & 34.4 \\
DynamicConv \citep{wu2018pay} & 35.2 \\
\midrule
Transformer (ours) & 34.70 \\
+ Word Dropout & 34.43 \\
+ RAML & 35.00 \\
+ SwitchOut & 35.28 \\
\midrule
+ \ssmba\ & \textbf{36.10\rlap{$^{*\dagger}$}} \\
\bottomrule
\end{tabular}
\caption{Results on IWSLT de$\to$en for models trained with different data augmentation methods. Scores marked with a $*$ and $\dagger$ are statistically significantly higher than baseline transformers and the next best model, both with $p<0.01$.}
\label{tab:iwslt_results}
\end{table}

\begin{table}[t!]
\small
\centering
\begin{tabular}{lcccc}
\toprule
& \multicolumn{2}{c}{\textbf{OPUS}} & \multicolumn{2}{c}{\textbf{de$\to$rm}} \\
\cmidrule(lr){2-3}
\cmidrule(lr){4-5}
Augmentation & ID & OOD & ID & OOD \\
\midrule
None & \textbf{56.99} & 10.24 & 51.53 & 12.23 \\
Word Dropout  & 56.26 & 10.15 & 50.23 & 12.23 \\
RAML & 56.76 & 10.10 & 51.52 & 12.49 \\
SwitchOut  & 55.50 & 9.27 & 51.34 & 13.59 \\
\midrule
\ssmba\  & 54.88 & \textbf{10.65} & \textbf{51.97} & \textbf{14.67\rlap{$^{*\dagger}$}} \\
\bottomrule
\end{tabular}
\caption{Average in-domain and out-of-domain BLEU for models trained on OPUS (de$\to$en) and de$\to$rm data. Scores marked with a $*$ and $\dagger$ are statistically significantly higher than baseline transformers and the next best model, both with $p<0.01$.}
\label{tab:nmt_results}
\end{table}

\subsection{Machine Translation}
\label{subsec:nmt_exp}
Table \ref{tab:iwslt_results} presents results on IWSLT14 de$\to$en.
We compare our results with convolutional models \citep{edunov-etal-2018-classical} and strong baseline transformer and dynamic convolution models \citep{wu2018pay}.
\ssmba\ improves BLEU by almost 1.5 points, outperforming all other baseline and comparison models.
Compared to \ssmba, other augmentation methods offer much smaller improvements or even degrade performance.

Table \ref{tab:nmt_results} presents results on OPUS and de$\to$rm.
On OPUS, where the training domain contains highly specialized language and differs significantly both from other domains and the learned MLM manifold, 
\ssmba\ offers a small boost in OOD BLEU but degrades ID performance. 
All other augmentation methods degrade both ID and OOD performance.
On de$\to$rm, \ssmba\ improves OOD BLEU by a large margin of 2.4 points, and ID BLEU by 0.4 points. 
Other augmentation methods offer much smaller OOD improvements while degrading ID performance.

\section{Analysis and Discussion}
\label{sec:discussion}
In this section, we analyze the factors that influence \ssmba's performance.
Due to its relatively small size (25k sentences), number of OOD domains (3), and amount of domain shift, we focus our analysis on the Baby domain within the AR-Clothing dataset.
Ablations are performed on a single domain rather than all domains, so error bars correspond to variance in models trained with different seeds and results are not comparable with those in Table \ref{tab:sent_results}.
Unless otherwise stated, we train CNN models and augment with \ssmba, corrupting 45\% of tokens, performing unrestricted sampling when reconstructing, and using self-supervised soft labelling, generating 5 synthetic examples for each training example. 

\subsection{Training Set Size}
\label{subsec:dsize_exp}
\begin{figure}[t]
\centering
\begin{tikzpicture}[scale=0.68, trim axis left]
\begin{axis}[
    xmode=log,
    log ticks with fixed point,
    xlabel=Training Set Size,
    ylabel=OOD Accuracy (\%),
    legend pos=north west,
    xtick pos=left,
    ytick pos=left,
    ymin=59,ymax=68,
    ymajorgrids=true,
    height=7.5cm, width=8.5cm,
]

\addplot+[blue, mark options={blue}, error bars/.cd, y dir=both, y explicit,] table [x=x, y=y,y error=error, col sep=comma] {
    x,  y,      error
    24, 59.65, 0.02
    98,	 59.75, 0.15
    390,	60.55, 0.47
    1562,	62.05, 0.67
    6250,  63.8,       0.58
    25000, 66.59, 0.76
};
\addlegendentry{No Augmentation}

\addplot+[red, mark options={red}, error bars/.cd, y dir=both, y explicit,] table [x=x, y=y,y error=error, col sep=comma] {
    x,  y,      error
24,	59.65, 0.06
98,	 59.92, 0.27
390,	61.06, 0.47
1562,	62.34, 0.56
6250,	64.26, 0.40
25000, 67.42, 0.41
};
\addlegendentry{\ssmba}

\comment{
\addplot table {
24 59.65
98	 59.75
390	60.55
1562	62.05
6250	63.8
25000 66.59
};
}

\end{axis}
\end{tikzpicture}
\caption{OOD accuracy of models trained on successively subsampled datasets. The full training set contains 25k examples. Error bars show standard deviation in OOD accuracy across models.}
\label{fig:dset_exp}
\end{figure}
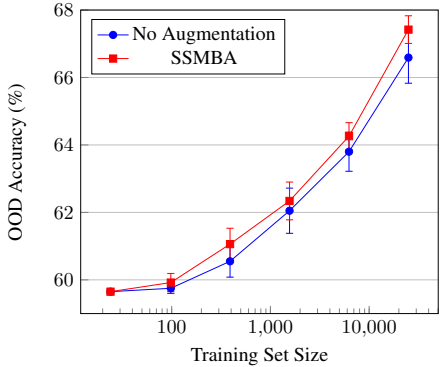

We first investigate how the size of the initial dataset affects \ssmba's effectiveness.
Since a smaller dataset covers less of the training distribution, we might expect the data generated by \ssmba\ to explore less of the data manifold and reduce its effectiveness.
We subsample 25\% of the original dataset to form a new training set, then repeat this process successively to form exponentially smaller and smaller datasets. 
The smallest dataset contains only 24 examples.
For each dataset fraction, we train 10 models and average performance,
tuning a set of \ssmba\ hyperparameters on the same ID validation data.
Figure \ref{fig:dset_exp} shows that \ssmba\ offers OOD performance gains across almost all dataset sizes, even in low resource settings with less than 100 training examples.

\subsection{Reconstruction Model Capacity}
\label{subsec:capacity_exp}
Since \ssmba\ relies on a reconstruction function that approximates the underlying data manifold, we might expect a larger and more expressive model to generate higher quality examples.
We investigate three models of varying size: DistilRoBERTa \citep{sanh2019distilbert} with 82M parameters, RoBERTa\textsubscript{BASE} with 125M parameters, and RoBERTa\textsubscript{LARGE} with 355M parameters. 
For each reconstruction model, we generate a set of 10 augmented datasets and train a set of 10 models on each augmented dataset. We average performance across models and datasests.
Table \ref{tab:recon_exp} shows that \ssmba\ displays robustness to the choice of reconstruction model, with all models conferring similar improvements to OOD accuracy. Using the smaller DistilRoBERTa model only degrades performance by a small margin.

\begin{table}[t]
    \small
    \centering
    \begin{tabular}{lccc}
        \toprule
        & Distil & Base & Large \\
        \midrule
        OOD Accuracy Boost (\%) & 0.73 & 0.78 & 0.78\\
        \bottomrule
    \end{tabular}
    \caption{Boost in OOD accuracy (\%) of models trained with \ssmba\ augmented data generated with different reconstruction functions.}
    \label{tab:recon_exp}
\end{table}

\comment{
\begin{table}[t]
    \small
    \centering
    \begin{tabular}{cc}
        \toprule
        & OOD Accuracy Boost (\%) \\
        \midrule
        Distil & 0.73 \\
        Base & 0.78 \\
        Large & 0.78 \\
        \bottomrule
    \end{tabular}
    \caption{Caption}
    \label{tab:my_label}
\end{table}
}

\comment{
\begin{figure}[t]
\centering
\begin{tikzpicture}[scale=0.7, trim axis left]
\begin{axis}[
    symbolic x coords={Distil, Base, Large},
    xtick=data, 
    bar width=45, 
    enlarge x limits=0.3,
    ymin=0,
    ylabel=Boost in OOD Accuracy (\%),
    xtick pos=left,
    ymajorgrids=true,
    ytick pos=left,
    height=7.5cm, width=8.5cm]
    
    \addplot[ybar,fill=blue, fill opacity=0.7,error bars/.cd, y dir=both, y explicit,] coordinates {
        (Distil, 0.73)
        (Base,0.78) 
        (Large,0.78)
    };
\end{axis}
\end{tikzpicture}
\caption{OOD accuracy of models trained with \ssmba\ augmented data generated with different reconstruction functions.
}
\label{fig:recon_exp}
\end{figure}
}

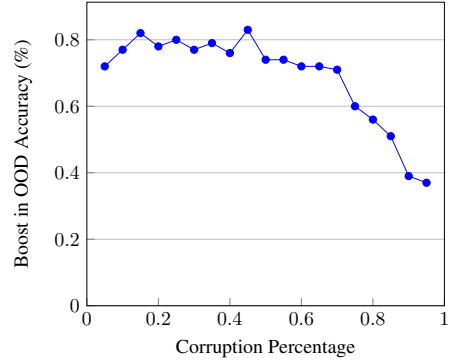
\begin{figure}[t]
\centering
\begin{tikzpicture}[scale=0.68, trim axis left]
\begin{axis}[
    xlabel=Corruption Percentage,
    ylabel=Boost in OOD Accuracy (\%),
    xtick pos=left,
    ytick pos=left,
    ymajorgrids=true,
    xmin=0,xmax=1,
    ymin=0,
    height=7.5cm, width=8.5cm,
]


\addplot+[blue, mark options={blue}] table [x=x, y=y,y error=error, col sep=comma] {
    x,  y,      error
0.05, 0.72, 0.45
0.1, 0.77, 0.39
0.15, 0.82, 0.38
0.2, 0.78, 0.44
0.25, 0.80, 0.39
0.3, 0.77, 0.37
0.35, 0.79, 0.42
0.4, 0.76, 0.37
0.45, 0.83, 0.42
0.5, 0.74, 0.39
0.55, 0.74, 0.41
0.6, 0.72, 0.41
0.65, 0.72, 0.41
0.7, 0.71, 0.42
0.75, 0.6, 0.42
0.8, 0.56, 0.42
0.85, 0.51, 0.45
0.9, 0.39, 0.47
0.95, 0.37, 0.47
};

\end{axis}
\end{tikzpicture}
\caption{Boost in OOD accuracy (\%) of models trained with \ssmba\ augmentation applied with different percentages of corrupted tokens.}
\label{fig:perturbation_exp}
\end{figure}

\subsection{Corruption Amount}
\label{subsec:noise_exp}
How sensitive is \ssmba\ to the particular amount of corruption applied?
Empirically,
tasks that were more sensitive to input noise, like sentiment analysis, required less corruption than those that were more robust, like NLI.
To analyze the effect of tuning the corruption amount, we generate 10 sets of augmented data with varying percentages of corruption, then train 10 models on each dataset, averaging performance across all 100 models.
Figure \ref{fig:perturbation_exp} shows that for corruption percentages below 50\%, our algorithm is relatively robust to the specific amount of corruption applied.
OOD performance peaks at 45\% corruption, decreasing thereafter as corruption increases.
Very large amounts of corruption tend to degrade performance, although surprisingly all augmented models still outperform unaugmented models, even when 95\% of tokens are corrupted.
In experiments on the more input sensitive NLI task, large amounts of noise degraded performance below baselines.

\subsection{Sample Generation Methods}
\label{subsec:sample_gen_exp}
Next we investigate methods for generating the reconstructed examples $\hat{x} \sim r(\hat{x}|x')$.
Top-k sampling draws samples from the MLM distribution on the top-k most probable tokens, leading to augmented data that explores higher probability regions of the manifold. 
We investigate top1, top5, top10, top20, and top50 sampling.
Unrestricted sampling draws samples from the full probability distribution of tokens.
This method explores a larger area of the underlying data distribution but can often lead to augmented data  in low probability regions.

\begin{figure}[t]
\centering
\begin{tikzpicture}[scale=0.68, trim axis left]
\begin{axis}[
    symbolic x coords={top1, top5, top10, top20, top50, unrestricted},
    xtick=data, 
    bar width=25, 
    xticklabel style={rotate=30,anchor=north east},
    ylabel=Boost in OOD Accuracy (\%),
    ymin=0,
    ymajorgrids=true,
    height=7.5cm, width=8.5cm,
    xtick pos=left,
    ytick pos=left,]
    
    \addplot[ybar, fill=blue, fill opacity=0.7,error bars/.cd, y dir=both, y explicit,] coordinates {
        (top1, 0.73) +- (0, 0.4)
        (top5, 0.72) +- (0, 0.4)
        (top10,0.62) +- (0, 0.38)
        (top20,0.7) +- (0, 0.42)
        (top50,0.7) +- (0, 0.4)
        (unrestricted,0.9) +- (0, 0.41)
    };
\end{axis}
\end{tikzpicture}
\caption{Boost in OOD accuracy (\%) of models trained with \ssmba\ augmentation using different sampling methods. Error bars show standard deviation in OOD accuracy across models.}
\label{fig:sample_gen_exp}
\end{figure}
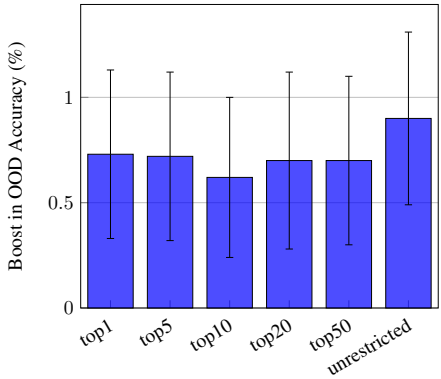

For each sample generation method, we generate 5 sets of augmented data and train 10 models on each dataset.
OOD accuracy is averaged across all models for a given sampling method.
Figure \ref{fig:sample_gen_exp} shows that unrestricted sampling provides the greatest increase in OOD accuracy, with top-k sampling methods all performing similarly.
This suggests that \ssmba\ works best when it is able to explore the manifold without any restrictions.

\subsection{Amount of Augmentation}
\label{subsec:aug_amount_exp}
\begin{figure}[t]
\centering
\begin{tikzpicture}[scale=0.68, trim axis left]
\begin{axis}[
    xlabel=\# Augmented Sentences,
    ylabel=OOD Accuracy (\%),
    legend pos=north east,
    xtick pos=left,
    ytick pos=left,
    xmin=-0.5, xmax=6.5,
    xticklabels={0,0,1,2,4,8,16,32},
    xtick distance=1,
    ymajorgrids=true,
    ymax=68,
    height=7.5cm, width=8.5cm,
]

\addplot[blue, mark=*, error bars/.cd, y dir=both, y explicit,] coordinates{
(0, 66.59) +- (0, 0.83)
(1, 66.77) +- (0, 0.54)
(2, 67.06) +- (0, 0.45)
(3, 67.31) +- (0, 0.36)
(4, 67.42) +- (0, 0.35)
(5, 67.53) +- (0, 0.29)
(6, 67.52) +- (0, 0.26)
};

\end{axis}
\end{tikzpicture}
\caption{OOD accuracy (\%) of models trained with different amounts of \ssmba\ augmentation. 0 augmentation corresponds to a baseline model. Error bars show standard deviation in OOD accuracy across models.}
\label{fig:num_exp}
\end{figure}
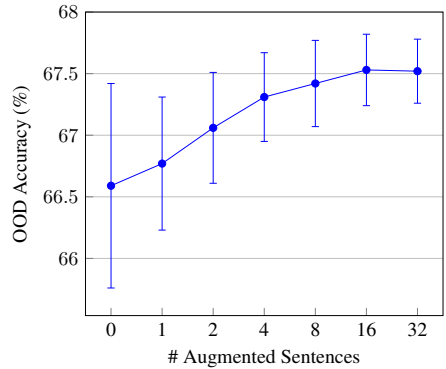

How does OOD accuracy change as we generate more sentences and explore more of the manifold neighborhood?
To investigate we select various augmentation amounts and generate 5 datasets for each amount, training 10 models on each dataset and averaging OOD accuracy across all 50 models.
Figure \ref{fig:num_exp} shows that increasing the amount of augmentation increases the amount by which \ssmba\ improves OOD accuracy, as well as decreasing the variance in the OOD accuracy of trained models.

\subsection{Label Generation}
\label{subsec:label_gen_exp}
We investigate 3 methods to generate a label $\hat{y}_{ij}$ for a synthetic example $\hat{x}_{ij}$.
\textit{Label preservation} preserves the original label $y_i$.
Since the manifold neighborhood of an example may cross a decision boundary, we also investigate using a supervised model $f$ trained on the original set of unaugmented data for
\textit{hard labelling} of a one-hot class label $\hat{y}_{ij}$ and \textit{soft labelling} of a class distribution $\hat{y}_{ij}$.

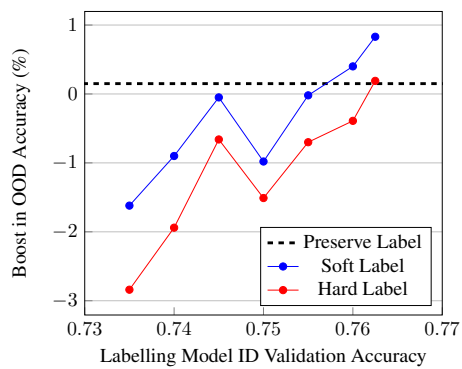
\begin{figure}[t]
\centering
\begin{tikzpicture}[scale=0.68, trim axis left]
\begin{axis}[
    xlabel=Labelling Model ID Validation Accuracy,
    ylabel=Boost in OOD Accuracy (\%),
    legend pos=south east,
    xtick pos=left,
    ytick pos=left,
    ymajorgrids=true,
    xmin=0.73,xmax=0.77,
    height=7.5cm, width=8.5cm,
]


\addplot[mark=none, dashed, ultra thick] coordinates {(0.72,0.15) (0.78,0.15)};
\addlegendentry{Preserve Label}

\addplot [blue, mark=*] table{
0.7625 0.83
0.76 0.4
0.755 -0.02
0.75 -0.98
0.745 -0.05
0.74 -0.9
0.735 -1.62
};
\addlegendentry{Soft Label}

\addplot [red, mark=*] table{
0.7625 0.19
0.76 -0.39
0.755 -0.7
0.75 -1.51
0.745 -0.66
0.74 -1.94
0.735 -2.84
};
\addlegendentry{Hard Label}

\end{axis}
\end{tikzpicture}
\caption{Boost in OOD accuracy (\%) of models trained with augmented data labelled with different supervision models and label generation methods.}
\label{fig:label_gen_exp}
\end{figure}

We train a CNN model to varying levels of convergence and validation accuracy, then label a set of 5 augmented datasets with each labelling method.
When training with soft labels, we optimize the KL-divergence between the output distribution and soft label distribution.
For each dataset we train 10 models and average performance across all models and datasets.
Results are shown in Figure \ref{fig:label_gen_exp}.

Unsurprisingly, soft and hard labelling with a low accuracy model degrades performance.
As our supervision classifier improves, so does the performance of models trained with soft and hard labelled data.
Once we pass a certain accuracy threshold, models trained with soft labels begin outperforming all other models.
This threshold varies depending on the difficulty of the dataset and task.
In ANLI experiments, labelling augmented examples even with a poor performing model still improved downstream accuracy.


\section{Conclusion}
\label{sec:conclusion}
In this paper, we introduce \ssmba, a method for generating synthetic data in settings where the underlying data manifold is difficult to characterize.
In contrast to other data augmentation methods, \ssmba\ is applicable to any supervised task, requires no task-specific knowledge, and does not rely on dataset-specific fine-tuning.
We demonstrate \ssmba's effectiveness on three NLP tasks spanning classification and sequence modeling: sentiment analysis, natural language inference, and machine translation.
We achieve gains of 0.8\% accuracy on OOD Amazon reviews, 1.8\% accuracy on OOD MNLI, and 1.4 BLEU on in-domain IWSLT14 de$\to$en.
Our analysis shows that \ssmba\ is robust to the initial dataset size, reconstruction model choice, and corruption amount, offering OOD robustness improvements in most settings.
Future work will explore applying \ssmba\ to the target side manifold in structured prediction tasks, as well as other natural language tasks and settings where data augmentation is difficult.

\section*{Acknowledgements}
Resources used in preparing this research were provided, in part, by the Province of Ontario, the Government of Canada through CIFAR, and companies sponsoring the Vector Institute \url{www.vectorinstitute.ai/#partners}.
This work was partly supported by Samsung Advanced Institute of Technology (Next Generation Deep Learning: from pattern recognition to AI) and Samsung Research (Improving Deep Learning using Latent Structure).
We thank Julian McAuley, Vishaal Prasad, Taylor Killian, Victoria Cheng, and Aparna Balagopalan for helpful comments and discussion.

\bibliography{anthology,emnlp2020}

\begin{thebibliography}{63}
\expandafter\ifx\csname natexlab\endcsname\relax\def\natexlab#1{#1}\fi

\bibitem[{Anaby-Tavor et~al.(2020)Anaby-Tavor, Carmeli, Goldbraich, Kantor,
  Kour, Shlomov, Tepper, and Zwerdling}]{lambada}
Ateret Anaby-Tavor, Boaz Carmeli, Esther Goldbraich, Amir Kantor, George Kour,
  Segev Shlomov, Naama Tepper, and Naama Zwerdling. 2020.
\newblock Do not have enough data? deep learning to the rescue!
\newblock In \emph{Proceedings of the 2020 AAAI}.

\bibitem[{Bachman et~al.(2014)Bachman, Alsharif, and
  Precup}]{bachman2014learning}
Philip Bachman, Ouais Alsharif, and Doina Precup. 2014.
\newblock Learning with pseudo-ensembles.
\newblock In \emph{NIPS}.

\bibitem[{Bengio et~al.(2013)Bengio, Yao, Alain, and
  Vincent}]{bengio2013generalized}
Yoshua Bengio, Li~Yao, Guillaume Alain, and Pascal Vincent. 2013.
\newblock Generalized denoising auto-encoders as generative models.
\newblock In \emph{Proceedings of the 26th International Conference on Neural
  Information Processing Systems - Volume 1}, NIPS’13, page 899–907, Red
  Hook, NY, USA. Curran Associates Inc.

\bibitem[{Blitzer et~al.(2007)Blitzer, Dredze, and
  Pereira}]{blitzer-etal-2007-biographies}
John Blitzer, Mark Dredze, and Fernando Pereira. 2007.
\newblock \href {https://www.aclweb.org/anthology/P07-1056} {Biographies,
  {B}ollywood, boom-boxes and blenders: Domain adaptation for sentiment
  classification}.
\newblock In \emph{Proceedings of the 45th Annual Meeting of the Association of
  Computational Linguistics}, pages 440--447, Prague, Czech Republic.
  Association for Computational Linguistics.

\bibitem[{Cettolo et~al.(2014)Cettolo, Niehues, St\"{u}ker, Bentivogli, and
  Federico}]{cettolo2014proceedings}
Mauro Cettolo, Jan Niehues, Sebastian St\"{u}ker, Luisa Bentivogli, and
  Marcello Federico. 2014.
\newblock Report on the 11th iwslt evaluation campaign, iwslt 2014.
\newblock In \emph{Proceedings of the 11$^{th}$ International Workshop on
  Spoken Language Translation}.

\bibitem[{Chan et~al.(2020)Chan, Möller, Pietsch, Soni, and
  Yeung}]{chan2020deepset}
Branden Chan, Timo Möller, Malte Pietsch, Tanay Soni, and Chin~Man Yeung.
  2020.
\newblock \href {https://deepset.ai/german-bert} {Open sourcing german bert}.

\bibitem[{Chapelle et~al.(2006)Chapelle, Schölkopf, and
  Zien}]{chapelle2006semi}
Olivier Chapelle, Bernhard Schölkopf, and Alexander Zien. 2006.
\newblock \emph{Semi-Supervised Learning (Adaptive Computation and Machine
  Learning)}.
\newblock The MIT Press.

\bibitem[{Chapelle et~al.(2000)Chapelle, Weston, Bottou, and
  Vapnik}]{vicinal200olivier}
Olivier Chapelle, Jason Weston, L\'{e}on Bottou, and Vladimir Vapnik. 2000.
\newblock Vicinal risk minimization.
\newblock In \emph{NIPS}.

\bibitem[{Daum{\'e}~III(2007)}]{daume-iii-2007-frustratingly}
Hal Daum{\'e}~III. 2007.
\newblock \href {https://www.aclweb.org/anthology/P07-1033} {Frustratingly easy
  domain adaptation}.
\newblock In \emph{Proceedings of the 45th Annual Meeting of the Association of
  Computational Linguistics}, pages 256--263, Prague, Czech Republic.
  Association for Computational Linguistics.

\bibitem[{Devlin et~al.(2018)Devlin, Chang, Lee, and Toutanova}]{devlin2018}
Jacob Devlin, Ming{-}Wei Chang, Kenton Lee, and Kristina Toutanova. 2018.
\newblock \href {http://arxiv.org/abs/1810.04805} {{BERT:} pre-training of deep
  bidirectional transformers for language understanding}.
\newblock \emph{CoRR}, abs/1810.04805.

\bibitem[{Edunov et~al.(2018)Edunov, Ott, Auli, Grangier, and
  Ranzato}]{edunov-etal-2018-classical}
Sergey Edunov, Myle Ott, Michael Auli, David Grangier, and Marc{'}Aurelio
  Ranzato. 2018.
\newblock \href {https://doi.org/10.18653/v1/N18-1033} {Classical structured
  prediction losses for sequence to sequence learning}.
\newblock In \emph{Proceedings of the 2018 Conference of the North {A}merican
  Chapter of the Association for Computational Linguistics: Human Language
  Technologies, Volume 1 (Long Papers)}, pages 355--364, New Orleans,
  Louisiana. Association for Computational Linguistics.

\bibitem[{Fadaee et~al.(2017)Fadaee, Bisazza, and Monz}]{fadaee2017data}
Marzieh Fadaee, Arianna Bisazza, and Christof Monz. 2017.
\newblock \href {https://doi.org/10.18653/v1/P17-2090} {Data augmentation for
  low-resource neural machine translation}.
\newblock In \emph{Proceedings of the 55th Annual Meeting of the Association
  for Computational Linguistics (Volume 2: Short Papers)}, pages 567--573,
  Vancouver, Canada. Association for Computational Linguistics.

\bibitem[{Hendrycks et~al.(2020)Hendrycks, Liu, Wallace, Dziedzic, Krishnan,
  and Song}]{hendrycks2020pretrained}
Dan Hendrycks, Xiaoyuan Liu, Eric Wallace, Adam Dziedzic, Rishabh Krishnan, and
  Dawn Song. 2020.
\newblock Pretrained transformers improve out-of-distribution robustness.
\newblock In \emph{Association for Computational Linguistics}.

\bibitem[{Hochreiter and Schmidhuber(1997)}]{hochreiter1997long}
Sepp Hochreiter and J{\"u}rgen Schmidhuber. 1997.
\newblock Long short-term memory.
\newblock \emph{Neural computation}, 9(8):1735--1780.

\bibitem[{Kafle et~al.(2017)Kafle, Yousefhussien, and
  Kanan}]{kafle-etal-2017-data}
Kushal Kafle, Mohammed Yousefhussien, and Christopher Kanan. 2017.
\newblock \href {https://doi.org/10.18653/v1/W17-3529} {Data augmentation for
  visual question answering}.
\newblock In \emph{Proceedings of the 10th International Conference on Natural
  Language Generation}, pages 198--202, Santiago de Compostela, Spain.
  Association for Computational Linguistics.

\bibitem[{Kanbak et~al.(2018)Kanbak, Seyed-Mohsen, and
  Frossard}]{kanbak2018geometric}
Can Kanbak, Moosavi-Dezfooli Seyed-Mohsen, and Pascal Frossard. 2018.
\newblock \href {https://doi.org/10.1109/CVPR.2018.00467} {Geometric robustness
  of deep networks: Analysis and improvement}.
\newblock pages 4441--4449.

\bibitem[{Kim(2014)}]{kim2014convolutional}
Yoon Kim. 2014.
\newblock Convolutional neural networks for sentence classification.
\newblock In \emph{Proceedings of the 2014 Conference on Empirical Methods in
  Natural Language Processing (EMNLP)}.

\bibitem[{Kingma and Ba(2014)}]{kingma2014method}
Diederik~P. Kingma and Jimmy Ba. 2014.
\newblock \href {http://arxiv.org/abs/1412.6980} {Adam: A method for stochastic
  optimization}.
\newblock Cite arxiv:1412.6980Comment: Published as a conference paper at the
  3rd International Conference for Learning Representations, San Diego, 2015.

\bibitem[{Kobayashi(2018)}]{kobayashi-2018-contextual}
Sosuke Kobayashi. 2018.
\newblock \href {https://doi.org/10.18653/v1/N18-2072} {Contextual
  augmentation: Data augmentation by words with paradigmatic relations}.
\newblock In \emph{Proceedings of the 2018 Conference of the North {A}merican
  Chapter of the Association for Computational Linguistics: Human Language
  Technologies, Volume 2 (Short Papers)}, pages 452--457, New Orleans,
  Louisiana. Association for Computational Linguistics.

\bibitem[{Koehn(2004)}]{koehn-2004-statistical}
Philipp Koehn. 2004.
\newblock \href {https://www.aclweb.org/anthology/W04-3250} {Statistical
  significance tests for machine translation evaluation}.
\newblock In \emph{Proceedings of the 2004 Conference on Empirical Methods in
  Natural Language Processing}, pages 388--395, Barcelona, Spain. Association
  for Computational Linguistics.

\bibitem[{Krizhevsky et~al.(2012)Krizhevsky, Sutskever, and
  Hinton}]{krizhevsky2012imagenet}
Alex Krizhevsky, Ilya Sutskever, and Geoffrey~E Hinton. 2012.
\newblock \href
  {http://papers.nips.cc/paper/4824-imagenet-classification-with-deep-convolutional-neural-networks.pdf}
  {Imagenet classification with deep convolutional neural networks}.
\newblock In F.~Pereira, C.~J.~C. Burges, L.~Bottou, and K.~Q. Weinberger,
  editors, \emph{Advances in Neural Information Processing Systems 25}, pages
  1097--1105. Curran Associates, Inc.

\bibitem[{{Kumar} et~al.(2020){Kumar}, {Choudhary}, and {Cho}}]{kumar20202data}
Varun {Kumar}, Ashutosh {Choudhary}, and Eunah {Cho}. 2020.
\newblock \href {http://arxiv.org/abs/2003.02245} {{Data Augmentation using
  Pre-trained Transformer Models}}.
\newblock \emph{arXiv e-prints}, page arXiv:2003.02245.

\bibitem[{Lewis et~al.(2019)Lewis, Liu, Goyal, Ghazvininejad, Mohamed, Levy,
  Stoyanov, and Zettlemoyer}]{lewis2019bart}
Mike Lewis, Yinhan Liu, Naman Goyal, Marjan Ghazvininejad, Abdelrahman Mohamed,
  Omer Levy, Veselin Stoyanov, and Luke Zettlemoyer. 2019.
\newblock Bart: Denoising sequence-to-sequence pre-training for natural
  language generation, translation, and comprehension.
\newblock \emph{arXiv preprint arXiv:1910.13461}.

\bibitem[{Liu et~al.(2019)Liu, Ott, Goyal, Du, Joshi, Chen, Levy, Lewis,
  Zettlemoyer, and Stoyanov}]{liu2019roberta}
Yinhan Liu, Myle Ott, Naman Goyal, Jingfei Du, Mandar Joshi, Danqi Chen, Omer
  Levy, Mike Lewis, Luke Zettlemoyer, and Veselin Stoyanov. 2019.
\newblock Roberta: A robustly optimized bert pretraining approach.
\newblock \emph{arXiv preprint arXiv:1907.11692}.

\bibitem[{Maas et~al.(2011)Maas, Daly, Pham, Huang, Ng, and
  Potts}]{maas2011learning}
Andrew~L. Maas, Raymond~E. Daly, Peter~T. Pham, Dan Huang, Andrew~Y. Ng, and
  Christopher Potts. 2011.
\newblock \href {http://www.aclweb.org/anthology/P11-1015} {Learning word
  vectors for sentiment analysis}.
\newblock In \emph{Proceedings of the 49th Annual Meeting of the Association
  for Computational Linguistics: Human Language Technologies}, pages 142--150,
  Portland, Oregon, USA. Association for Computational Linguistics.

\bibitem[{Miyato et~al.(2017)Miyato, Maeda, Koyama, and
  Ishii}]{miyato2017virtual}
Takeru Miyato, Shin-ichi Maeda, Masanori Koyama, and Shin Ishii. 2017.
\newblock \href {http://arxiv.org/abs/1704.03976} {{Virtual Adversarial
  Training: A Regularization Method for Supervised and Semi-Supervised
  Learning}}.

\bibitem[{Müller et~al.(2019)Müller, Gonzales, and
  Sennrich}]{Muller2019DomainRI}
Mathias Müller, Annette~Rios Gonzales, and Rico Sennrich. 2019.
\newblock Domain robustness in neural machine translation.
\newblock \emph{ArXiv}, abs/1911.03109.

\bibitem[{Ni et~al.(2019)Ni, Li, and McAuley}]{jianmo}
Jianmo Ni, Jiacheng Li, and Julian McAuley. 2019.
\newblock Justifying recommendations using distantly-labeled reviews and
  fined-grained aspects.
\newblock In \emph{Proceedings of EMNLP}.

\bibitem[{Nie et~al.(2019)Nie, Williams, Dinan, Bansal, Weston, and
  Kiela}]{nie2019adversarial}
Yixin Nie, Adina Williams, Emily Dinan, Mohit Bansal, Jason Weston, and Douwe
  Kiela. 2019.
\newblock \href {http://arxiv.org/abs/1910.14599} {{Adversarial NLI: A New
  Benchmark for Natural Language Understanding}}.

\bibitem[{Norouzi et~al.(2016)Norouzi, Bengio, Chen, Jaitly, Schuster, Wu, and
  Schuurmans}]{norouzi2016reward}
Mohammad Norouzi, Samy Bengio, Zhifeng Chen, Navdeep Jaitly, Mike Schuster,
  Yonghui Wu, and Dale Schuurmans. 2016.
\newblock \href
  {http://papers.nips.cc/paper/6547-reward-augmented-maximum-likelihood-for-neural-structured-prediction.pdf}
  {Reward augmented maximum likelihood for neural structured prediction}.
\newblock In D.~D. Lee, M.~Sugiyama, U.~V. Luxburg, I.~Guyon, and R.~Garnett,
  editors, \emph{Advances in Neural Information Processing Systems 29}, pages
  1723--1731. Curran Associates, Inc.

\bibitem[{Ott et~al.(2019)Ott, Edunov, Baevski, Fan, Gross, Ng, Grangier, and
  Auli}]{ott2019fairseq}
Myle Ott, Sergey Edunov, Alexei Baevski, Angela Fan, Sam Gross, Nathan Ng,
  David Grangier, and Michael Auli. 2019.
\newblock fairseq: A fast, extensible toolkit for sequence modeling.
\newblock In \emph{Proceedings of NAACL-HLT 2019: Demonstrations}.

\bibitem[{Papineni et~al.(2002)Papineni, Roukos, Ward, and
  Zhu}]{papineni2002bleu}
Kishore Papineni, Salim Roukos, Todd Ward, and Wei-Jing Zhu. 2002.
\newblock \href {https://doi.org/10.3115/1073083.1073135} {Bleu: A method for
  automatic evaluation of machine translation}.
\newblock In \emph{Proceedings of the 40th Annual Meeting on Association for
  Computational Linguistics}, ACL ’02, page 311–318, USA. Association for
  Computational Linguistics.

\bibitem[{Paschali et~al.(2019)Paschali, Simson, Roy, Naeem, G\"{o}bl,
  Wachinger, and Navab}]{paschali2019data}
Magdalini Paschali, Walter Simson, Abhijit~Guha Roy, Muhammad~Ferjad Naeem,
  R\"{u}diger G\"{o}bl, Christian Wachinger, and Nassir Navab. 2019.
\newblock Data augmentation with manifold exploring geometric transformations
  for increased performance and robustness.
\newblock \emph{arXiv}.

\bibitem[{Perez and Wang(2017)}]{perez2017}
Luis Perez and Jason Wang. 2017.
\newblock \href {http://arxiv.org/abs/1712.04621} {{The Effectiveness of Data
  Augmentation in Image Classification using Deep Learning}}.

\bibitem[{Post(2018)}]{post-2018-call}
Matt Post. 2018.
\newblock \href {https://doi.org/10.18653/v1/W18-6319} {A call for clarity in
  reporting {BLEU} scores}.
\newblock In \emph{Proceedings of the Third Conference on Machine Translation:
  Research Papers}, pages 186--191, Belgium, Brussels. Association for
  Computational Linguistics.

\bibitem[{Quionero-Candela et~al.(2009)Quionero-Candela, Sugiyama,
  Schwaighofer, and Lawrence}]{quionera2009dataset}
Joaquin Quionero-Candela, Masashi Sugiyama, Anton Schwaighofer, and Neil~D.
  Lawrence. 2009.
\newblock \emph{Dataset Shift in Machine Learning}.

\bibitem[{Rico~Sennrich(2016)}]{sennrich2016improving}
Alexandra~Birch Rico~Sennrich, Barry~Haddow. 2016.
\newblock Improving neural machine translation models with monolingual data.
\newblock In \emph{Proc. of ACL}.

\bibitem[{Sajjadi et~al.(2016)Sajjadi, Javanmardi, and
  Tasdizen}]{sajjadi2016regularization}
Mehdi Sajjadi, Mehran Javanmardi, and Tolga Tasdizen. 2016.
\newblock Regularization with stochastic transformations and perturbations for
  deep semi-supervised learning.
\newblock In \emph{NIPS}.

\bibitem[{Sanh et~al.(2019)Sanh, Debut, Chaumond, and
  Wolf}]{sanh2019distilbert}
Victor Sanh, Lysandre Debut, Julien Chaumond, and Thomas Wolf. 2019.
\newblock Distilbert, a distilled version of bert: smaller, faster, cheaper and
  lighter.
\newblock In \emph{NeurIPS $EMC^2$ Workshop}.

\bibitem[{Scherrer and Cartoni(2012)}]{scherrer-cartoni-2012-trilingual}
Yves Scherrer and Bruno Cartoni. 2012.
\newblock \href
  {http://www.lrec-conf.org/proceedings/lrec2012/pdf/685_Paper.pdf} {The
  trilingual {ALLEGRA} corpus: Presentation and possible use for lexicon
  induction}.
\newblock In \emph{Proceedings of the Eighth International Conference on
  Language Resources and Evaluation ({LREC}-2012)}, pages 2890--2896, Istanbul,
  Turkey. European Languages Resources Association (ELRA).

\bibitem[{Sennrich et~al.(2016)Sennrich, Haddow, and
  Birch}]{sennrich-etal-2016-edinburgh}
Rico Sennrich, Barry Haddow, and Alexandra Birch. 2016.
\newblock \href {https://doi.org/10.18653/v1/W16-2323} {{E}dinburgh neural
  machine translation systems for {WMT} 16}.
\newblock In \emph{Proceedings of the First Conference on Machine Translation:
  Volume 2, Shared Task Papers}, pages 371--376, Berlin, Germany. Association
  for Computational Linguistics.

\bibitem[{Simard et~al.(1998)Simard, LeCun, Denker, and Victorri}]{Simard1998}
Patrice~Y. Simard, Yann~A. LeCun, John~S. Denker, and Bernard Victorri. 1998.
\newblock \href {https://doi.org/10.1007/3-540-49430-8_13}
  {\emph{Transformation Invariance in Pattern Recognition --- Tangent Distance
  and Tangent Propagation}}, pages 239--274. Springer Berlin Heidelberg,
  Berlin, Heidelberg.

\bibitem[{Simonyan and Zisserman(2015)}]{simonyan2014very}
Karen Simonyan and Andrew Zisserman. 2015.
\newblock Very deep convolutional networks for large-scale image recognition.
\newblock In \emph{Proceedings of the 2015 International Conference on Learning
  Representations}.

\bibitem[{Socher et~al.(2013)Socher, Perelygin, Wu, Chuang, Manning, Ng, and
  Potts}]{socher2013recursive}
Richard Socher, Alex Perelygin, Jean Wu, Jason Chuang, Christopher~D. Manning,
  Andrew Ng, and Christopher Potts. 2013.
\newblock \href {https://www.aclweb.org/anthology/D13-1170} {Recursive deep
  models for semantic compositionality over a sentiment treebank}.
\newblock In \emph{Proceedings of the 2013 Conference on Empirical Methods in
  Natural Language Processing}, pages 1631--1642, Seattle, Washington, USA.
  Association for Computational Linguistics.

\bibitem[{Szegedy et~al.(2014)Szegedy, Zaremba, Sutskever, Bruna, Erhan,
  Goodfellow, and Fergus}]{szegedy2014intriguing}
Christian Szegedy, Wojciech Zaremba, Ilya Sutskever, Joan Bruna, Dumitru Erhan,
  Ian Goodfellow, and Rob Fergus. 2014.
\newblock \href {http://arxiv.org/abs/1312.6199} {Intriguing properties of
  neural networks}.
\newblock In \emph{International Conference on Learning Representations}.

\bibitem[{Tiedemann(2012)}]{TIEDEMANN12.463}
J\"{o}rg Tiedemann. 2012.
\newblock Parallel data, tools and interfaces in opus.
\newblock In \emph{Proceedings of the Eight International Conference on
  Language Resources and Evaluation (LREC'12)}, Istanbul, Turkey. European
  Language Resources Association (ELRA).

\bibitem[{{Torralba} and {Efros}(2011)}]{torralba2011unbiased}
A.~{Torralba} and A.~A. {Efros}. 2011.
\newblock Unbiased look at dataset bias.
\newblock In \emph{CVPR 2011}, pages 1521--1528.

\bibitem[{Vaswani et~al.(2017)Vaswani, Shazeer, Parmar, Uszkoreit, Jones,
  Gomez, Kaiser, and Polosukhin}]{vaswani2017attention}
Ashish Vaswani, Noam Shazeer, Niki Parmar, Jakob Uszkoreit, Llion Jones,
  Aidan~N. Gomez, Lukasz Kaiser, and Illia Polosukhin. 2017.
\newblock Attention is all you need.
\newblock In \emph{Proceedings of the 2017 Conference on Neural Information
  Processing Systems}.

\bibitem[{Vincent et~al.(2008)Vincent, Larochelle, Bengio, and
  Manzagol}]{vincent2008extracting}
Pascal Vincent, Hugo Larochelle, Yoshua Bengio, and Pierre-Antoine Manzagol.
  2008.
\newblock Extracting and composing robust features with denoising autoencoders.
\newblock In \emph{Proceedings of the 25th International Conference on Machine
  Learning}.

\bibitem[{Wang and Yang(2015)}]{wangyang2015thats}
William~Yang Wang and Diyi Yang. 2015.
\newblock \href {https://doi.org/10.18653/v1/D15-1306} {That{'}s so
  annoying!!!: A lexical and frame-semantic embedding based data augmentation
  approach to automatic categorization of annoying behaviors using {\#}petpeeve
  tweets}.
\newblock In \emph{Proceedings of the 2015 Conference on Empirical Methods in
  Natural Language Processing}, pages 2557--2563, Lisbon, Portugal. Association
  for Computational Linguistics.

\bibitem[{Wang et~al.(2018)Wang, Pham, Dai, and
  Neubig}]{wang-etal-2018-switchout}
Xinyi Wang, Hieu Pham, Zihang Dai, and Graham Neubig. 2018.
\newblock \href {https://doi.org/10.18653/v1/D18-1100} {{S}witch{O}ut: an
  efficient data augmentation algorithm for neural machine translation}.
\newblock In \emph{Proceedings of the 2018 Conference on Empirical Methods in
  Natural Language Processing}, pages 856--861, Brussels, Belgium. Association
  for Computational Linguistics.

\bibitem[{Wei and Zou(2019)}]{wei-zou-2019-eda}
Jason Wei and Kai Zou. 2019.
\newblock \href {https://www.aclweb.org/anthology/D19-1670} {{EDA}: Easy data
  augmentation techniques for boosting performance on text classification
  tasks}.
\newblock In \emph{Proceedings of the 2019 Conference on Empirical Methods in
  Natural Language Processing and the 9th International Joint Conference on
  Natural Language Processing (EMNLP-IJCNLP)}, pages 6383--6389, Hong Kong,
  China. Association for Computational Linguistics.

\bibitem[{Williams et~al.(2018)Williams, Nangia, and
  Bowman}]{williams2018broad}
Adina Williams, Nikita Nangia, and Samuel Bowman. 2018.
\newblock \href {http://aclweb.org/anthology/N18-1101} {A broad-coverage
  challenge corpus for sentence understanding through inference}.
\newblock In \emph{Proceedings of the 2018 Conference of the North American
  Chapter of the Association for Computational Linguistics: Human Language
  Technologies, Volume 1 (Long Papers)}, pages 1112--1122. Association for
  Computational Linguistics.

\bibitem[{Wu et~al.(2019{\natexlab{a}})Wu, Fan, Baevski, Dauphin, and
  Auli}]{wu2018pay}
Felix Wu, Angela Fan, Alexei Baevski, Yann Dauphin, and Michael Auli.
  2019{\natexlab{a}}.
\newblock \href {https://arxiv.org/abs/1901.10430} {Pay less attention with
  lightweight and dynamic convolutions}.
\newblock In \emph{International Conference on Learning Representations}.

\bibitem[{Wu et~al.(2019{\natexlab{b}})Wu, Lv, Zang, Han, and
  Hu}]{wu2019conditional}
Xing Wu, Shangwen Lv, Liangjun Zang, Jizhong Han, and Songlin Hu.
  2019{\natexlab{b}}.
\newblock Conditional bert contextual augmentation.
\newblock In \emph{International Conference on Computational Science}, pages
  84--95. Springer.

\bibitem[{Xie et~al.(2019)Xie, Dai, Hovy, Luong, and Le}]{xie2019unsupervised}
Qizhe Xie, Zihang Dai, Eduard Hovy, Minh-Thang Luong, and Quoc~V Le. 2019.
\newblock Unsupervised data augmentation for consistency training.
\newblock \emph{arXiv preprint arXiv:1904.12848}.

\bibitem[{Xie et~al.(2020)Xie, Dai, Hovy, Luong, and Le}]{xie2020unsupervised}
Qizhe Xie, Zihang Dai, Eduard Hovy, Minh-Thang Luong, and Quoc~V. Le. 2020.
\newblock \href {https://openreview.net/forum?id=ByeL1R4FvS} {Unsupervised data
  augmentation for consistency training}.

\bibitem[{Xie et~al.(2017)Xie, Wang, Li, Levy, Nie, Jurafksy, and
  Ng}]{xie2017data}
Ziang Xie, Sida~I. Wang, Jiwei Li, Daniel Levy, Aiming Nie, Dan Jurafksy, and
  Andrew~Y. Ng. 2017.
\newblock Data noising as smoothing in neural network language models.
\newblock In \emph{Proceedings of the 2017 International Conference on Learning
  Representations}.

\bibitem[{{Yang} et~al.(2020){Yang}, {Malaviya}, {Fernandez}, {Swayamdipta},
  {Le Bras}, {Wang}, {Bhagavatula}, {Choi}, and {Downey}}]{yang2020g-daug}
Yiben {Yang}, Chaitanya {Malaviya}, Jared {Fernandez}, Swabha {Swayamdipta},
  Ronan {Le Bras}, Ji-Ping {Wang}, Chandra {Bhagavatula}, Yejin {Choi}, and
  Doug {Downey}. 2020.
\newblock \href {http://arxiv.org/abs/2004.11546} {{G-DAUG: Generative Data
  Augmentation for Commonsense Reasoning}}.
\newblock \emph{arXiv e-prints}, page arXiv:2004.11546.

\bibitem[{Yelp()}]{yelp}
Yelp.
\newblock Yelp open dataset.
\newblock \url{https://www.yelp.com/dataset}.

\bibitem[{Yu et~al.(2018)Yu, Dohan, Le, Luong, Zhao, and Chen}]{wei2018fast}
Adams~Wei Yu, David Dohan, Quoc Le, Thang Luong, Rui Zhao, and Kai Chen. 2018.
\newblock \href {https://openreview.net/forum?id=B14TlG-RW} {Fast and accurate
  reading comprehension by combining self-attention and convolution}.
\newblock In \emph{International Conference on Learning Representations}.

\bibitem[{Zhang et~al.(2018)Zhang, Cisse, Dauphin, and
  Lopez-Paz}]{zhang2018mixup}
Hongyi Zhang, Moustapha Cisse, Yann~N. Dauphin, and David Lopez-Paz. 2018.
\newblock \href {https://openreview.net/forum?id=r1Ddp1-Rb} {mixup: Beyond
  empirical risk minimization}.
\newblock \emph{International Conference on Learning Representations}.

\bibitem[{Zhang et~al.(2015)Zhang, Zhao, and LeCun}]{zhang2015character}
Xiang Zhang, Junbo Zhao, and Yann LeCun. 2015.
\newblock \href
  {http://papers.nips.cc/paper/5782-character-level-convolutional-networks-for-text-classification.pdf}
  {Character-level convolutional networks for text classification}.
\newblock In C.~Cortes, N.~D. Lawrence, D.~D. Lee, M.~Sugiyama, and R.~Garnett,
  editors, \emph{Advances in Neural Information Processing Systems 28}, pages
  649--657. Curran Associates, Inc.

\end{thebibliography}
\bibliographystyle{acl_natbib}

\clearpage
\appendix
\section{Datasets}
\label{app:data}
Full dataset statistics and details are provided in table \ref{tab:data_summ_big}.
All data splits for all tasks can be downloaded at \url{https://nyu.box.com/s/henvmy17tkyr6npl7e1ltw8j46baxsml}. 

\begin{table*}[t]
\small
\centering
\begin{tabular}{llcccccc}
\toprule
Dataset & Domain & Reference & $c$ & $l$ & Train & Valid & Test \\
\midrule
\multirow{4}{*}{AR-Clothing}
& Men & \citealt{jianmo}  & 5 & 31 & 25k$^\dagger$ & 2k & 2k\\
& Women & \citealt{jianmo}  & 5 & 40 & 25k$^\dagger$ & 2k & 2k\\
& Baby & \citealt{jianmo} & 5 & 29 & 25k$^\dagger$ & 2k & 2k\\
& Shoes & \citealt{jianmo} & 5 & 41 & 25k$^\dagger$ & 2k & 2k\\
\midrule
\multirow{10}{*}{AR-Full}
& Books & \citealt{jianmo} & 5 & 101 & 25k$^\dagger$ & 2k & 2k\\
& Clothing, Shoes \& Jewelry & \citealt{jianmo} & 5 & 39 & 25k$^\dagger$ & 2k & 2k\\
& Home and Kitchen & \citealt{jianmo} & 5 & 53 & 25k$^\dagger$ & 2k & 2k\\
& Kindle Store & \citealt{jianmo} & 5 & 104 & 25k$^\dagger$ & 2k & 2k\\
& Movies \& TV & \citealt{jianmo} & 5 & 83 & 25k$^\dagger$ & 2k & 2k\\
& Pet Supplies & \citealt{jianmo} & 5 & 57 & 25k$^\dagger$ & 2k & 2k\\
& Sports \& Outdoors & \citealt{jianmo} & 5 & 55 & 25k$^\dagger$ & 2k & 2k\\
& Electronics & \citealt{jianmo} & 5 & 73 & 25k$^\dagger$ & 2k & 2k\\
& Tools \& Home Improvement & \citealt{jianmo} & 5 & 57 & 25k$^\dagger$ & 2k & 2k\\
& Toys \& Games & \citealt{jianmo} & 5 & 50 & 25k$^\dagger$ & 2k & 2k\\
\midrule
\multirow{4}{*}{Yelp} 
& American & \citealt{yelp} & 5 & 138 & 25k$^\dagger$ & 2k & 2k \\
& Chinese & \citealt{yelp} & 5 & 135 & 25k$^\dagger$ & 2k & 2k\\
& Italian & \citealt{yelp} & 5 & 139 & 25k$^\dagger$ & 2k & 2k\\
& Japanese & \citealt{yelp} & 5 & 138 & 25k$^\dagger$ & 2k & 2k\\
\midrule
\multirow{10}{*}{MNLI}
& Slate & \citealt{williams2018broad} & 3 & 35 & 75k & 2k & 2k \\
& Fiction & \citealt{williams2018broad} & 3 & 25 & 73k & 2k & 2k \\
& Telephone & \citealt{williams2018broad} & 3 & 37 & 81k & 2k & 2k \\
& Travel & \citealt{williams2018broad} & 3 & 42 & 75k& 2k & 2k  \\
& Government & \citealt{williams2018broad} & 3 & 39 & 75k & 2k & 2k  \\
& Verbatim & \citealt{williams2018broad} & 3 & 43 & - & 1k & 1k \\
& Face-to-Face & \citealt{williams2018broad} & 3 & 29 & - & 1k & 1k \\
& OUP & \citealt{williams2018broad} & 3 & 41 & - & 1k & 1k \\
& 9/11 & \citealt{williams2018broad} & 3 & 36 & - & 1k & 1k \\
& Letters & \citealt{williams2018broad} & 3 & 34 & - & 1k & 1k \\
\midrule
\multirow{2}{*}{Movies}
& SST2 & \citealt{socher2013recursive} & 2 & 11 & 66k & 1k & 1k\\
& IMDb & \citealt{maas2011learning} & 2 & 296 & 46k & 2k & 2k\\
\midrule
\multirow{3}{*}{ANLI}
& R1 & \citealt{nie2019adversarial} & 3 & 92 & 17k & 1k & 1k \\
& R2 & \citealt{nie2019adversarial} & 3 & 90 & 46k & 1k & 1k \\
& R3 & \citealt{nie2019adversarial} & 3 & 82 & 100k & 1k & 1k \\
\midrule
IWSLT & IWSLT & \citealt{cettolo2014proceedings} & - & 24 & 160k & 7k & 7k \\
\midrule
\multirow{5}{*}{OPUS}
& Medical & \citealt{TIEDEMANN12.463} & - & 13 & 1.1m & 2k & 2k\\
& IT & \citealt{TIEDEMANN12.463} & - & 14 & - & 2k & 2k\\
& Koran & \citealt{TIEDEMANN12.463} & - & 23 & - & 2k & 2k\\
& Law & \citealt{TIEDEMANN12.463} & - & 31 & - & 2k & 2k\\
& Subtitles & \citealt{TIEDEMANN12.463} & - & 10 & - & 2k & 2k\\
\midrule
\multirow{2}{*}{de$\to$rm}
& Law & \citealt{scherrer-cartoni-2012-trilingual} & - & 22 & 101k & 2k & 2k \\
& Blogs & \citealt{Muller2019DomainRI} & - & 24 & - & 2k & 2k \\
\bottomrule
\end{tabular}
\caption{Summary statistics for datasets. For detailed information, see references. $n$: number of domains. $c$: number of target classes. $l$: average training example length, or average test example length, for datasets without training sets. Training sets marked with a $\dagger$ are sampled randomly from a larger dataset.}
\label{tab:data_summ_big}
\end{table*}

\section{Data Preprocessing}
\label{app:preprocess}
We use the same preprocessing steps across all sentiment analysis and NLI experiments.
All data is first tokenized using a GPT-2 style tokenizer and BPE vocabulary provided by \texttt{fairseq} \citep{ott2019fairseq}.
This BPE vocabulary consists of 50263 types. 
Corresponding labels are encoded using a label dictionary consisting of as many types as there are classes.
Input text and labels are then binarized for model training. 
Although all models share the same vocabulary, we randomly initialize each model's embeddings and train the entire model end-to-end.
For machine translation experiments, we follow \citealt{Muller2019DomainRI} and learn a 16k BPE on OPUS and a 32k BPE on de$\to$rm. 
On IWSLT14 we learn a 10k BPE.
We use a separate vocabulary for the source and target side.

\section{Model Architecture and Training Hyperparameters}
\label{app:hyperparams}

All models are written and trained within the \texttt{fairseq} framework \citep{ott2019fairseq} with T4 GPUs.
LSTM and CNN models were trained on a single GPU, RoBERTa models were trained with 4 GPUs, and tranfsormer models were trained with 2 GPUs. 
On average, when trained on augmented data, LSTM and CNN models took an hour to train to convergence, RoBERTa models took 12 hours to train to convergence, and transformer models took 24 hours to train to convergence. 
Models trained on unaugmented data took roughly 20\% of the time of models trained on augmented data to reach convergence. 
For each model we investigate, we present first the model architecture and then the training hyperparameters.

\subsection{LSTM}
\label{subapp:lstms}
Our LSTM models are a single layer of 512 nodes. 
Input embeddings are 512 dimensions.
The output embedding from the last time step is fed into a MLP classifier with a single hidden layer of 512 dimensions.
Models contain 28M parameters.
Dropout of 0.3 is applied to the input and output of our encoder, and dropout of 0.1 is applied to the MLP classifier.

We train with Adam optimizer \citep{kingma2014method} with $\beta = (0.9, 0.98)$ and $\epsilon =$ \num{1e-6}.
Our learning rate is set to \num{1e-4} and is first warmed up for 2 epochs before it is decayed using an inverse square root scheduler.

\subsection{CNN}
\label{subapp:cnn}
Our CNN models are based on the architecture in \cite{kim2014convolutional}. 
As in our LSTM models, our input embeddings are 512 dimensional, which we treat as our channel dimension.
We apply three convolutions of kernel size 3, 4, and 5, with 256 output channels. 
Models contain 27M parameters.
Convolutional outputs are max-pooled over time then concatenated to a 768-dimensional encoded representation.
Again, we feed this representation into a MLP classifier with a single hidden layer of 512 dimensions.
We apply dropout of 0.2 to our inputs and MLP classifier.

We train with Adam optimizer \citep{kingma2014method} with $\beta = (0.9, 0.98)$ and $\epsilon =$ \num{1e-6}.
Our learning rate is set to \num{1e-3} and is first warmed up for 2 epochs before it is decayed using an inverse square root scheduler.

\subsection{RoBERTa}
\label{subapp:roberta}
Our RoBERTa models use a pre-trained RoBERTa\textsubscript{BASE} model provided by \texttt{fairseq}. 
As in other models, classification token embeddings are fed into an MLP classifier with a single hidden layer of 512 dimensions.
Models contain 125M parameters.
We follow the MNLI fine-tuning procedures in \texttt{fairseq}, training with learning rate \num{1e-5} with Adam optimizer \citep{kingma2014method} with $\beta = (0.9, 0.98)$ and $\epsilon =$ \num{1e-6}.
We warmup the learning rate for 2 epochs before decaying with an inverse square root scheduler.

\subsection{Transformer}
\label{subapp:transformer}
Transformer models are trained with label-smoothed cross-entropy and label smoothing $0.1$. 
Due to the dataset sizes, we use a slightly smaller transformer architecture with embedding dimension 512, feed forward embedding dimension 1024, 4 encoder heads, and 6 encoder and decoder layers.
Models contain 52M parameters.
We also apply dropout of 0.3 and weight decay of 0.0001.
All other hyperparameters follow the base architecture in \citealt{vaswani2017attention}.

As in other models, we train with Adam optimizer \citep{kingma2014method} with $\beta = (0.9, 0.98)$ and $\epsilon =$ \num{1e-6}.
Our learning rate is set to \num{5e-4} and is first warmed up for 4000 updates before it is decayed using an inverse square root scheduler.

\section{\ssmba\ Hyperparameters}
\label{app:ssmba}
\ssmba\ hyperparameters for each dataset and domain are provided in table \ref{tab:ssmba_hyperparams}. 
Hyperparameters are chosen based on in-domain validation performance. A detailed analysis of hyperparameter tuning is provided in section \ref{sec:discussion}.

\begin{table*}[t]
\small
\centering
\begin{tabular}{llccccc}
\toprule
Dataset & Domain & Model & Corruption \% & Sampling Method & Labelling Method & \# Generated\\
\midrule
\multirow{2}{*}{AR-Clothing}
& * & RNN & 40\% & Unrestricted Sampling & Preserve Label & 5\\
& * & CNN & 40\% & Unrestricted Sampling & Soft Label & 5\\
\midrule
\multirow{2}{*}{AR-Full}
& * & RNN & 50\% & Unrestricted Sampling & Preserve Label & 5\\
& * & CNN & 40\% & Unrestricted Sampling & Soft Label & 5\\
\midrule
\multirow{2}{*}{Yelp}
& * & RNN & 60\% & Unrestricted Sampling & Preserve Label & 5\\
& * & CNN & 40\% & Unrestricted Sampling & Soft Label & 5\\
\multirow{4}{*}{Movies}
& SST2 & RNN & 10\% & Unrestricted Sampling & Soft Label & 5\\
& IMDb & RNN & 20\% & Unrestricted Sampling & Preserve Label & 5\\
& SST2 & CNN & 60\% & Unrestricted Sampling & Hard Label & 5\\
& IMDb & CNN & 30\% & Unrestricted Sampling & Soft Label & 5\\
\midrule
MNLI & * & RoBERTa & 10\% & Unrestricted Sampling & Soft Label & 5\\
\midrule
\multirow{3}{*}{ANLI}
& R1 & RoBERTa & 5\% & Unrestricted Sampling & Preserve Label & 5\\
& R2 & RoBERTa & 5\% & Unrestricted Sampling & Hard Label & 5\\
& R3 & RoBERTa & 10\% & Unrestricted Sampling & Hard Label & 5\\
\midrule
IWSLT & IWSLT & Transformer & 10\% & Unrestricted Sampling & Beam 5 & 5\\
\midrule
OPUS & Medical & Transformer & 15\% & Unrestricted Sampling & Beam 5 & 5\\
\midrule
de$\to$rm & Law & Transformer & 15\% & Unrestricted Sampling & Beam 5 & 5\\
\bottomrule
\end{tabular}
\caption{\ssmba\ hyperparameters used to generate augmented data for each dataset and domain. Hyperparameters were selected by in-domain validation performance. A * in the domain indicates that hyperparameters are the same for all domains in that dataset.}
\label{tab:ssmba_hyperparams}
\end{table*}

\section{Statistical Testing}
\label{app:stats}
For the statistical tests on sentiment analysis and NLI tasks, we use a Wilcoxon ranked-sum test. 
Specifically, we compare averages of model performances on pairs of training and test domains.
For example, in a dataset with 3 domains, D1, D2, and D3, we have 3 in-domain train-test pairs (D1-D1, D2-D2, D3-D3), and 6 out-of-domain train-test pairs (D1-D2, D1-D3, D2-D1, D2-D3, D3-D1, D3-D2).
We calculate the average performance for each model on each pair, then compare the matched in-domain and out-of-domain pairs. 
Since the number of samples we can compare depends on the total number of domains in the dataset, a larger number of datasets gives us a better sense of our statistical significance.

For the statistical tests on machine translation tasks, we use a paired bootstrap resampling approach \citep{koehn-2004-statistical}. 
Since the test works only on a single system's output, we run the test on every pairing of seeds and test domains for the two comparison models.
We report the significance level only if all tests result in a small enough probability.

\end{document}